\documentclass[runningheads]{llncs}
\usepackage[T1]{fontenc}
\usepackage[utf8]{inputenc}
\usepackage{graphicx}
\usepackage{amsmath}        
\usepackage{amssymb}        
\usepackage{amsfonts}       
\usepackage{booktabs}       
\usepackage{array}          
\usepackage{multirow}       
\usepackage{caption}        
\usepackage{float}          
\usepackage{bm}             
\usepackage{enumitem}       
\usepackage{siunitx}          
\usepackage[table]{xcolor}
\usepackage[most]{tcolorbox}

\definecolor{lightblue}{RGB}{217, 234, 249}
\usepackage{fontawesome5}  
\usepackage{adjustbox}
\usepackage{float}
\usepackage{tabularx}
\usepackage{algorithm, algpseudocode}
\usepackage{enumitem}
\usepackage{booktabs}
\usepackage{makecell}
\usepackage{tabularx}
\usepackage{hyperref}
\usepackage{orcidlink}
\algrenewcommand{\alglinenumber}[1]{\scriptsize\arabic{ALG@line}}

\begin{document}
\title{A Transformer Based Handwriting Recognition System Jointly Using Online and Offline Features}
\titlerunning{Joint Utilisation of Online and Offline Features for Handwriting Recognition}

\author{%
  Ayush Lodh\inst{1}\orcidlink{0009-0001-7506-0900}%
    \thanks{Ayush Lodh and Ritabrata Chakraborty contributed equally to this work.}%
  \and
  Ritabrata Chakraborty\inst{1,2}\orcidlink{0009-0009-3597-3703}%
    \thanks{Work done during internship at Indian Statistical Institute.}%
  \and
  Shivakumara Palaiahnakote\inst{3}\orcidlink{0000-0001-9026-4613}%
  \and
  Umapada Pal\inst{1}\orcidlink{0000-0002-5426-2618}%
}

\institute{%
  Indian Statistical Institute, India\\
  \email{\{ayushlodh26,\,ritabrata04\}@gmail.com}\\
  \email{umapada@isical.ac.in}
  \and
  Manipal University Jaipur, India
  \and
  University of Salford, United Kingdom\\
  \email{s.palaiahnakote@salford.ac.uk}
}

\authorrunning{A. Lodh, R. Chakraborty et al.}
\maketitle              

\begin{abstract}
We posit that handwriting recognition benefits from complementary cues carried by the rasterized complex glyph and the pen’s trajectory, yet most systems exploit only one modality. We introduce an end-to-end network that performs early fusion of offline images and online stroke data within a shared latent space. A patch encoder converts the grayscale crop into fixed-length visual tokens, while a lightweight transformer embeds the $(x,y,\mathrm{pen})$ sequence. Learnable latent queries attend jointly to both token streams, yielding context-enhanced stroke embeddings that are pooled and decoded under a cross-entropy loss objective. Because integration occurs before any high-level classification, temporal cues reinforce each other during representation learning, producing stronger writer independence. Comprehensive experiments on IAMOn-DB, and VNOn-DB demonstrate that our approach achieves state-of-the-art accuracy, exceeding previous bests by up to 1\%.  Our study also shows adaptation of this pipeline with gesturification on the ISI-Air dataset. Our code can be found  \href{https://github.com/AZTECLUPR/HATChar-Classifier}{here.} 
\end{abstract}

\section{Introduction}
Handwritten Text Recognition (HTR) is a foundational task in the broader domain of handwriting analysis \cite{jimaging10010018}, with widespread applications in document digitization, education technology, and human-computer interaction. While many advances have been made in recognizing isolated characters using deep learning, most existing models are trained on datasets such as UNIPEN \cite{guyon1994unipen}, comprising neatly segmented, individual characters written in isolation. However, this is far removed from real-world scenarios, where characters look different when embedded within a word, influenced by neighboring characters, and written in a variety of natural handwriting styles.
\begin{figure}[H]
    \centering
    \includegraphics[width=1\linewidth]{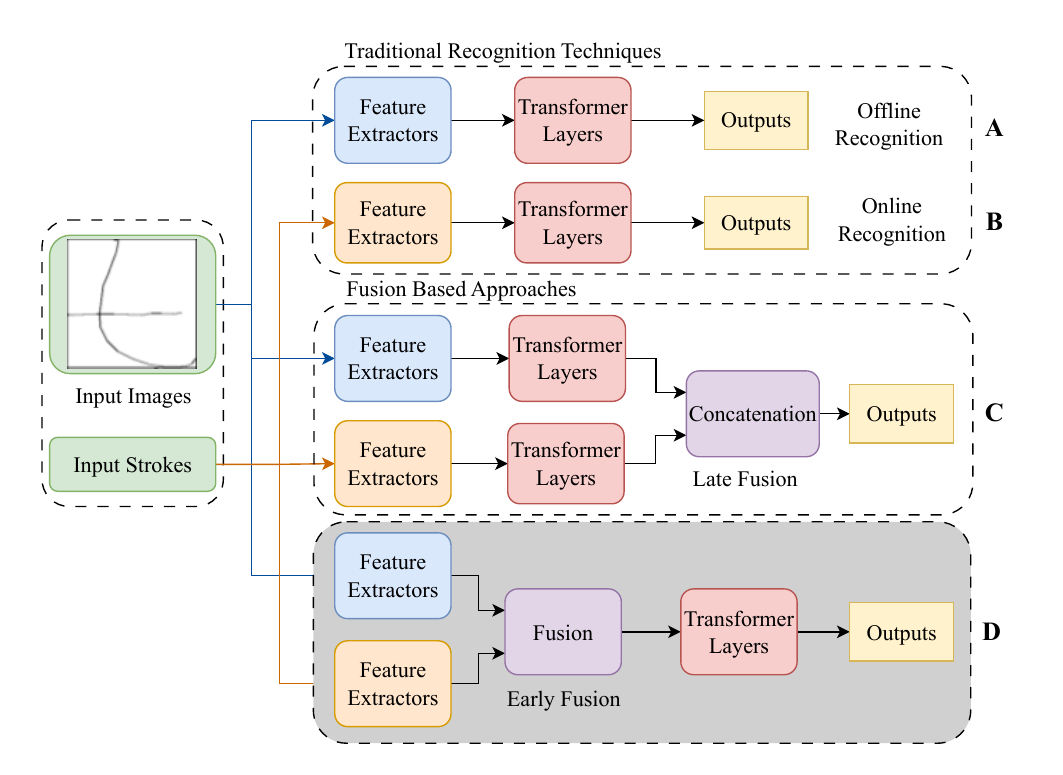}
    \caption{An overview of input modalities and their architectures for handwritten text recognition. A: Image-only input, B: Stroke-only input, C: Late Fusion based dual input, D: Early Fusion based dual input (Ours).}
    \label{fig:types-of-recognition}
\end{figure}
\vspace{-0.5cm}
Early handwritten‐character recognition systems were designed around a single modality. Offline pipelines (Fig. \ref{fig:types-of-recognition}-A) treat a scanned glyph as an image and rely on convolutional or transformer backbones to map pixels to character sequences \cite{li2025htr}. Conversely, online pipelines (Fig. \ref{fig:types-of-recognition}-B) discard appearance altogether, interpreting only the pen-tip trajectory captured by the digitiser \cite{digitizer}. Each view is incomplete as images lose temporal order, while stroke traces omit shading, pen pressure, and context such as ligatures or serifs. 
To mitigate this limitation, bimodal fusion emerged in a multiscale network \cite{xu2024multi} that concatenates high-level image and stroke embeddings before classification (late fusion; Fig. \ref{fig:types-of-recognition}-C). Such strategies boost robustness, yet they still process both streams independently until the penultimate layer. As a result, misaligned timelines, redundant features, and modality-specific noise remain unaddressed, and the network cannot learn shared primitives—e.g., a cusp that appears as both a sharp curvature in trajectory space and a dark pixel cluster in image space.

We posit that early fusion (Fig. \ref{fig:types-of-recognition}-D) unlocks deeper complementarity. \textbf{By projecting raw visual patches and stroke tokens into a shared latent space and allowing cross-attention before any task-specific transformer layers, the model can utilise the interaction between how a character looks in pixel space and how it is traced through the pen vectors.} We utilize IAMOn-DB \cite{liwicki2005iam} and VNOn-DB \cite{nguyen2018icfhr} datasets adapted for character-level recognition to deisgn our experimental study for this hypothesis. We also display recognition results on the ISI-Air \cite{rahman2021air} dataset, showing strong results to air-writing data.

\begin{enumerate}[label=\roman*.]
    \item We explore a novel direction in the field of handwritten text recognition, leveraging the relationship of character images and the pen stroke vectors used in tracing them.
    \item We propose HATCharClassifier, a first of its kind handwritten text recognizer that provides robust character recognition results through early fusion of multiple inputs (image + stroke).
    \item We benchmark our model as state-of-the-art for multiple multimodal handwritten text datasets such as IAMOn-DB~\cite{liwicki2005iam} and VNOn-DB~\cite{nguyen2018icfhr}, along with air writing datasets like ISI-Air~\cite{rahman2021air}.
\end{enumerate}

The rest of the paper is structured as follows: Section 2 explores related literature on early classical methods, deep learning and transformer based methods, and fusion based models for handwritten text recognition. Section 3 explores our proposed framework HATCharClassifier. Section 4 describes experiment design, datasets used and metrics we employ to show the performance of our method. Section 5 delves into quantitative and qualitative results for the proposed framework. Section 6 provides a discussion on the importance of such a method with regards to current research and its caveats, and finally, Section 7 concludes the paper.

\section{Related Work}
\subsection{Early Methods for Handwriting Recognition}
Statistical and classical machine learning approaches have played a pivotal role in advancing handwritten text recognition in the early years of the field. Hidden Markov Models (HMMs) emerged as the dominant framework for modeling sequential handwriting, particularly excelling in cursive script recognition due to their ability to perform implicit segmentation and probabilistic modeling of character sequences \cite{bozinovic1989off,plamondon2000online,marti2002iam,bengio1995lerec}. Support Vector Machines (SVMs) gained traction for isolated character recognition by leveraging margin-based classification and kernel methods, with adaptations such as dynamic time warping kernels enabling their application to online handwriting sequences \cite{bahlmann2002online}. To capture richer dependencies, discriminative models like Conditional Random Fields (CRFs) were also explored, for example,\cite{feng2006exploring} showed that CRFs can outperform HMMs on whole-word recognition tasks.  These classical methods typically operate on carefully engineered features.  For instance, Bai and Huo \cite{bai2005study} extract 8-directional histogram features from pen trajectories for online Chinese character recognition.  Likewise, many systems convert raw strokes into offline images or other local descriptors to feed into the sequence model. Alongside these, k-Nearest Neighbors (k-NN) and shallow neural networks like multilayer perceptrons served as competitive baselines for digit and character recognition, especially on benchmarks such as UNIPEN \cite{guyon1994unipen} and CEDAR \cite{cedarcdrom-1}. These classifiers were highly reliant on carefully engineered features such as zoning, projection histograms, contour profiles, and geometric descriptors—extracted from normalized and preprocessed handwriting samples \cite{impedovo2014zoning,plamondon2000online}.  Together, these classical methods laid the algorithmic foundation for contemporary handwritten text recognition systems.

\subsection{Deep Learning Based Methods}
With the advent of deep learning \cite{lecun2015deep}, end-to-end neural models became standard for HTR. \cite{shi2016end} introduced the CRNN architecture, combining CNN feature extraction with bidirectional RNN sequence modeling.  Bi-directional LSTM (BiLSTM) networks \cite{huang2015bidirectional} or gated RNNs then capture long-range context in the stroke/image sequence. \cite{graves2008novel} demonstrate that a BLSTM trained with Connectionist Temporal Classification (CTC) loss \cite{graves2006connectionist} can significantly outperform traditional HMM baselines on unconstrained handwriting recognition.  Encoder–decoder models with attention have also been applied for lexicon-free transcription. More recently, fully Transformer-based OCR models have appeared. For example, \cite{li2023trocr} propose TrOCR, which uses a pre-trained Vision Transformer encoder and text Transformer decoder, yielding state-of-the-art results on handwritten text recognition benchmarks. These works paved the way towards performance-maximising models suitable for multilingual and multidomain use-cases.

\subsection{Recent Attention-Based Methods}
With the advent of attention mechanisms \cite{vaswani2017attention}, transformer architectures have begun to be applied to handwriting recognition. \cite{li2025htr} adapted the Vision Transformer (ViT) \cite{dosovitskiy2020image} for line-level text recognition. Their HTR-VT model uses a CNN for feature extraction and employs sharpness-aware minimization, achieving competitive accuracy on standard HTR datasets like IAMOn-DB~\cite{liwicki2005iam}. \cite{jungo2023character} introduce Character Queries, a transformer decoder where each character is represented by a learned query vector; this approach excels at segmenting on-line strokes into characters given a known transcription. C-TST \cite{chen2023online}, a two-stream model using a 1D convolution + Transformer branch for temporal stroke features and a Vision Transformer for spatial image features; fusing both streams yields high accuracy on Chinese benchmarks. \cite{li2025chinese} utilized the Swin Transformer as the encoder to extract image features, focusing on Chinese character characteristics. 
These Transformer-based methods complement and often improve upon earlier RNN and CNN-based systems.

\subsection{Bimodal Fusion Methods}
Multimodal (image + stroke) fusion has been widely studied to improve robustness.  Most recent methods employ late-fusion multi-stream architectures: separate encoders process pen trajectories and images, and their outputs are merged.  For example, \cite{xu2024multi} propose a multi-scale bimodal fusion network that combines features from both streams using Transformers, achieving state-of-the-art accuracy on IAMOn-DB (e.g.\ 4.7\% CER).
Similarly, Bhunia et al. ~\cite{bhunia2020indic} fuse online trajectory features with rendered images for Indic script recognition.  While these late-fusion models yield high accuracy, they incur extra complexity due to separate image rendering and fusion modules.  We identify this as a \textbf{domain gap}: training stroke and image encoders independently can limit joint feature learning.  To address this, we introduce an \emph{early fusion} strategy that jointly embeds stroke and image information from the outset.

\section{Methodology}\label{sec: methodology}
\vspace{-1cm}
\begin{figure}[h]
    \centering
    \includegraphics[width=1\linewidth]{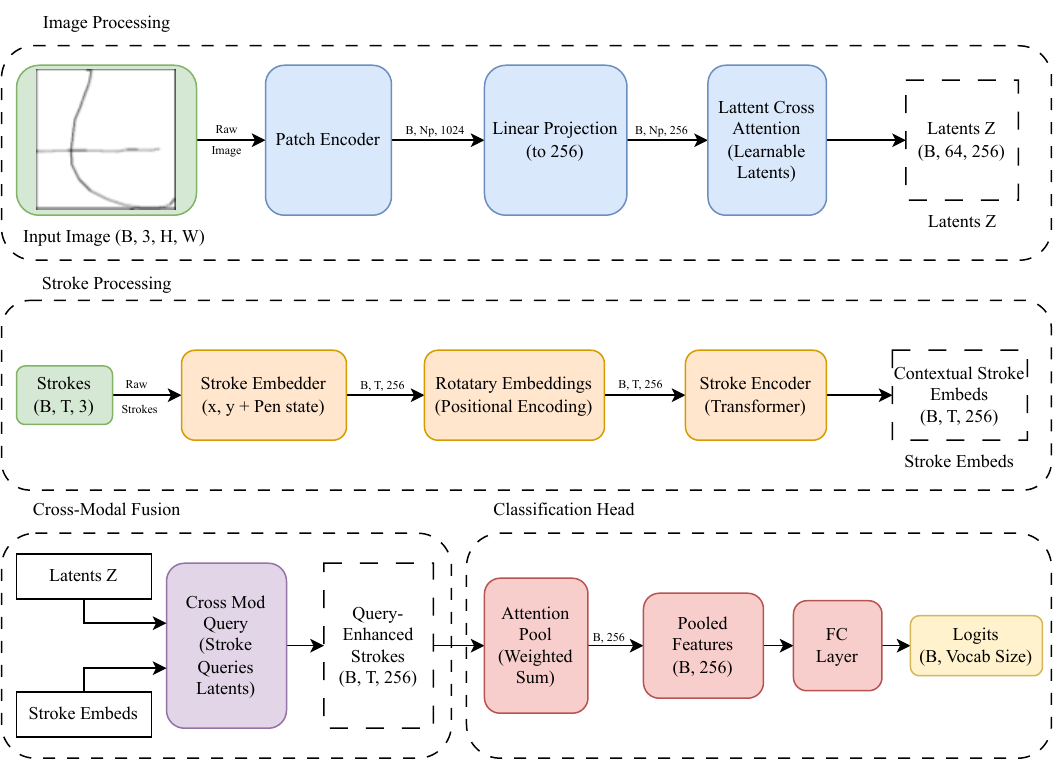}
    \caption{Our proposed pipeline.}
    \label{fig:model_architecture}
\end{figure}

\label{sec:hat}
\vspace{-1cm}
\paragraph{\textbf{Notation.}}
Let
$\mathcal D=\{(\mathbf I_i,\mathbf S_i,y_i)\}_{i=1}^{M}$ with  
$\mathbf I_i\!\in\!\mathbb R^{H\times W\times C}$ (gray or RGB image),  
$\mathbf S_i=[(x_t,y_t,p_t)]_{t=1}^{T_i}\!\in\!\mathbb R^{T_i\times3}$ (online
stroke sequence, $p_t\!\in\!\{0,1\}$ pen state), and  
$y_i\!\in\!\{1,\dots,V\}$.  
HAT converts either modality—or both—into a common $d$-dimensional token
space and classifies with a linear head (Fig.~\ref{fig:model_architecture}).
\paragraph{\textbf{Image Patch Encoder.}}
\label{sec:image-enc}

We first (bi-linearly) resize $\mathbf I$ to $224\times224$ and, when
necessary, replicate its single channel to obtain $C{=}3$.  A pretrained
Swin-B~\cite{liu2021swin} backbone outputs the last-stage feature map
$\mathbf F\!\in\!\mathbb R^{7\times7\times1024}$.  Flattening the spatial
axes and projecting with $\mathbf W_p\!\in\!\mathbb R^{1024\times d}$ gives
\begin{equation}
  \mathbf E_p
  \;=\;
  \operatorname{reshape}(\mathbf F)\,\mathbf W_p
  \in\mathbb R^{N\times d},
  \qquad N=7\times7 .
  \label{eq:patch-proj}
\end{equation}

\paragraph{\textbf{Latent Cross-Attention.}}
\label{sec:latent}

Following Perceiver-IO~\cite{jaegle2021perceiver}, we introduce
$L$ learnable \emph{latent tokens} $\mathbf Z^{(0)}\!\in\!\mathbb R^{L\times d}$.
For layers $\ell=0,\dots,L_{\text{img}}-1$
\begin{subequations}
\begin{align}
  \tilde{\mathbf Z}^{(\ell)} &=
    \mathrm{MHA}(\mathbf Q{=}\mathbf Z^{(\ell)},
                 \mathbf K{=}\mathbf E_p,
                 \mathbf V{=}\mathbf E_p),
                 \label{eq:latent-mha}\\
  \mathbf Z^{(\ell+1)} &=
    \mathrm{TransformerLayer}\!\bigl(
      \mathbf Z^{(\ell)} + \tilde{\mathbf Z}^{(\ell)}
    \bigr).
    \label{eq:latent-tl}
\end{align}
\end{subequations}

Layer-norm on the final state yields
$\mathbf Z=\mathrm{LN}(\mathbf Z^{(L_{\text{img}})})$.

\paragraph{\textbf{Stroke Encoder.}}
\label{sec:stroke-enc}
\vspace{-0.3cm}
Each point $(x_t,y_t,p_t)$ is embedded by concatenating raw coordinates with
a pen-state lookup
$\mathbf E_{\text{pen}}\!\in\!\mathbb R^{2\times d/8}$:
\begin{equation}
  \mathbf s_t=[x_t,y_t,\mathbf E_{\text{pen}}[p_t]]
  \in\mathbb R^{2+d/8}.
\end{equation}
After a point-wise projection, BatchNorm, and Dropout we obtain
\begin{equation}
  \mathbf E_s =
  \bigl[\mathbf s_t\bigr]_{t=1}^{T}\mathbf W_s
  \in\mathbb R^{T\times d},
  \qquad
  \mathbf W_s\!\in\!\mathbb R^{(2+d/8)\times d}.
  \label{eq:stroke-proj}
\end{equation}

\paragraph{\textbf{Rotary positional encoding.}}
Splitting every token into even/odd parts and rotating with angle
$\phi_t=t\boldsymbol\Theta$ (see \cite{su2024roformer}) gives
$\hat{\mathbf E}_s\!\in\!\mathbb R^{T\times d}$.

\paragraph{\textbf{Temporal transformer.}}
An $N_{\text{stk}}$-layer Transformer processes the sequence
\begin{equation}
  \mathbf H=\mathrm{TransformerEncoder}(\hat{\mathbf E}_s)
  \in\mathbb R^{T\times d},
  \label{eq:stroke-transformer}
\end{equation}
which is refined via a residual 2-layer MLP:
\begin{equation}
  \mathbf E_{\text{stroke}} =
    \mathbf H + \mathrm{MLP}\!\bigl(\mathrm{LN}(\mathbf H)\bigr).
  \label{eq:stroke-final}
\end{equation}

\paragraph{\textbf{Cross-Modal Querying.}}
\label{sec:crossmod}

With both modalities present, stroke tokens query the latent image tokens:
\begin{subequations}
\begin{align}
  \tilde{\mathbf E}_{\text{stk}} &=
    \mathrm{MHA}(\mathbf Q{=}\mathbf E_{\text{stroke}},
                 \mathbf K{=}\mathbf Z,
                 \mathbf V{=}\mathbf Z),            \\
  \mathbf E_{\text{cross}} &=
    \mathrm{TransformerLayer}\!\bigl(
      \mathbf E_{\text{stroke}}+\tilde{\mathbf E}_{\text{stk}}
    \bigr).
    \label{eq:cross-update}
\end{align}
\end{subequations}
If images (resp.\ strokes) are missing we set
$\mathbf T=\mathbf Z$ (resp.\ $\mathbf T=\mathbf E_{\text{stroke}}$).

\paragraph{\textbf{Attention Pooling \& Classification.}}
\label{sec:pool}

Given token matrix $\mathbf T\!\in\!\mathbb R^{N_t\times d}$
($N_t=L$ or $T$), scalar importances
\begin{equation}
  \alpha_i =
  \frac{\exp\!\bigl(
    \mathbf w_2^{\!\top}\tanh(\mathbf W_1\mathbf t_i)
  \bigr)}
  {\sum_{j=1}^{N_t}
   \exp\!\bigl(
    \mathbf w_2^{\!\top}\tanh(\mathbf W_1\mathbf t_j)
  \bigr)},
  \qquad
  \mathbf g=\sum_{i=1}^{N_t}\alpha_i\mathbf t_i ,
  \label{eq:attn-pool}
\end{equation}
are computed with $\mathbf W_1\!\in\!\mathbb R^{d\times d}$,
$\mathbf w_2\!\in\!\mathbb R^{d}$.  The classifier is
\begin{equation}
  \mathbf o=\mathbf W_c\mathbf g+\mathbf b_c
  \in\mathbb R^{V},
  \label{eq:classifier}
\end{equation}
and we minimise cross-entropy
\begin{equation}
  \mathcal L=
  -\frac1M\sum_{i=1}^{M}
   \log\frac{\exp(o_{i,y_i})}{\sum_{v=1}^{V}\exp(o_{i,v})}.
  \label{eq:loss}
\end{equation}
\vspace{-1cm}

\begin{algorithm}[H]
  \caption{\textbf{HAT} 
           (refer to Eq.~\eqref{eq:patch-proj}–\eqref{eq:loss}as aforementioned)}
  \label{alg:hat_training}
  \begin{algorithmic}[1]
    \Require mini-batch $\{(\mathbf I_i,\mathbf S_i,y_i)\}_{i=1}^{B}$,
             mode $m\!\in\!\{\textsc{Image},\textsc{Stroke},\textsc{Both}\}$
    \For{$i\gets1$ \textbf{to} $B$}
      \If{$m\neq\textsc{Stroke}$} \Comment{image branch
            }
        \State $\mathbf E_p\gets\textsc{ImagePatchEncoder}(\mathbf I_i)$
               \Comment{Eq.~\eqref{eq:patch-proj}}
        \State $\mathbf Z\gets\textsc{LatentCrossAttn}(\mathbf E_p)$
               \Comment{Eqs.~\eqref{eq:latent-mha}–\eqref{eq:latent-tl}}
      \EndIf
      \If{$m\neq\textsc{Image}$} \Comment{stroke branch:
            Sec.~\ref{sec:stroke-enc}}
        \State $\mathbf E_{\text{stk}}\gets\textsc{StrokeEncoder}(\mathbf S_i)$
               \Comment{Eqs.~\eqref{eq:stroke-proj}–\eqref{eq:stroke-final}}
      \EndIf
      \State \textbf{Select} $\mathbf T$
        \Statex\hspace{\algorithmicindent}\textbullet~
          $m{=}\textsc{Image}$  $\!\!\rightarrow\!\mathbf T\!=\!\mathbf Z$
        \Statex\hspace{\algorithmicindent}\textbullet~
          $m{=}\textsc{Stroke}$ $\!\!\rightarrow\!\mathbf T\!=\!
                                   \mathbf E_{\text{stk}}$
      \If{$m=\textsc{Both}$}      \Comment{hybrid:
            Sec.~\ref{sec:crossmod}}
        \State $\mathbf T\gets\textsc{CrossModalQuery}(
                     \mathbf E_{\text{stk}},\mathbf Z)$
               \Comment{Eq.~\eqref{eq:cross-update}}
      \EndIf
      \State $\mathbf g\gets\textsc{AttentionPool}(\mathbf T)$
             \Comment{Eq.~\eqref{eq:attn-pool}}
      \State $\mathbf o_i\gets\mathbf W_c\mathbf g+\mathbf b_c$
             \Comment{Eq.~\eqref{eq:classifier}}
    \EndFor
    \State $\mathcal L\gets
           \textsc{CrossEntropy}(\{\mathbf o_i\},\{y_i\})$
           \Comment{Eq.~\eqref{eq:loss}}
    \State \textbf{Back-propagate} $\nabla\mathcal L$; update parameters
  \end{algorithmic}
\end{algorithm}

\vspace{-0.5cm}
\begin{tcolorbox}[colback=yellow!10,  
                  colframe=yellow!40!black, 
                  boxrule=0.5pt,
                  arc=2pt,
                  left=4pt,right=4pt,top=4pt,bottom=4pt]
\textbf{In summary,} HAT maps images (via a frozen Swin\textsubscript{B}) and online strokes (via a rotary-encoded transformer) to a shared $d$-dimensional token space. Stroke tokens optionally query image tokens through a single cross-modal attention layer, after which attentional pooling yields a global vector for linear classification with cross-entropy. The same lightweight architecture handles image-only, stroke-only, and hybrid inputs without altering parameters.
\end{tcolorbox}
\vspace{-0.5cm}

\section{Experiments}

\subsection{Datasets}
\label{sec:data}
\vspace{-0.5cm}
\begin{figure*}[h]
  \centering
  \begin{minipage}[t]{0.50\textwidth}
    \captionsetup{type=table}
    \captionof{table}{Dataset statistics after pre-processing}
    \label{tab:dataset_stats}
    \scriptsize                
    \setlength\tabcolsep{3pt}  
    \renewcommand\arraystretch{1.05}

    \begin{adjustbox}{max width=\textwidth}
      \begin{tabular}{
          l                     
          S[table-format=6.0]   
          S[table-format=6.0]   
          S[table-format=6.0]   
          S[table-format=4.0]   
        }
        \toprule
        \textbf{Dataset} &
        {\makecell{\textbf{Train}\\\textbf{Set}}} &
        {\makecell{\textbf{Val.}\\\textbf{Set}}} &
        {\makecell{\textbf{Test}\\\textbf{Set}}} &
        {\makecell{\textbf{\#}\\\textbf{Cls.}}} \\
        \midrule
        IAMOn-DB~\cite{liwicki2005iam} & {72,508} & {18,954} & {21,455} &  {84} \\
        VNOn-DB~\cite{nguyen2018icfhr} & {197,140} & {57,763} & {77,389} & {145} \\
        ISI-Air~\cite{rahman2021air}   & {10,000} &  {2,000} & {-}  & {10} \\
        \bottomrule
      \end{tabular}
    \end{adjustbox}

    \vspace{3pt}
    \footnotesize Numbers = character samples.
  \end{minipage}%
  \hfill
  \begin{minipage}[t]{0.46\textwidth}
    \vspace{0pt} 
    We evaluate on three on-line handwriting corpora, each trained and
    tested in isolation. Qualitative examples are given in
    Fig.~\ref{fig:iam_examples},~\ref{fig:vnondb_examples},~\ref{fig:isi_examples},
    while Table~\ref{tab:dataset_stats} (left) reports the final number
    of character instances per split together with the number of target
    classes.
  \end{minipage}
\end{figure*}

\vspace{-0.5cm}
The \textbf{IAMOn-DB} \cite{liwicki2005iam} dataset is a widely used benchmark for online handwriting recognition, particularly focused on English cursive script. It contains handwritten text samples collected from 221 writers using a stylus on a tablet, capturing the temporal sequence of pen strokes along with their spatial coordinates. IAMOn-DB supports writer-independent and writer-dependent evaluation protocols and is frequently used for training and evaluating sequence models like HMMs and RNNs. Its high-quality online handwriting data has made it a standard in evaluating temporal modeling capabilities in handwriting recognition systems.
    \vspace{-0.5cm}
    \begin{figure}[h]
    \centering
    \includegraphics[width=1\linewidth]{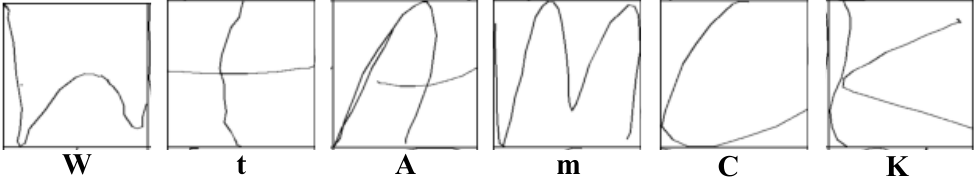}
    \vspace{-0.8cm}
    \caption{Some examples from IAMOn-DB dataset.}
    \label{fig:iam_examples}
\end{figure}

\vspace{-0.5cm}

The VNOn-DB (Vietnamese Online Handwriting Database) \cite{nguyen2018icfhr} is a large-scale dataset designed to support research in Vietnamese online handwriting recognition. It comprises pen trajectory data collected from over 200 writers, covering all 134 Vietnamese characters including diacritics. Each character is annotated with stroke order and pen-up/pen-down events, providing rich temporal and spatial information. VNOn-DB presents challenges specific to the Vietnamese language, such as compound characters and tonal marks, making it a valuable resource for evaluating script-specific handwriting models. The dataset has been used to benchmark both character-level and word-level recognition tasks in low-resource language settings.
\vspace{-0.5cm}
    \begin{figure}[h]
    \centering
    \includegraphics[width=1\linewidth]{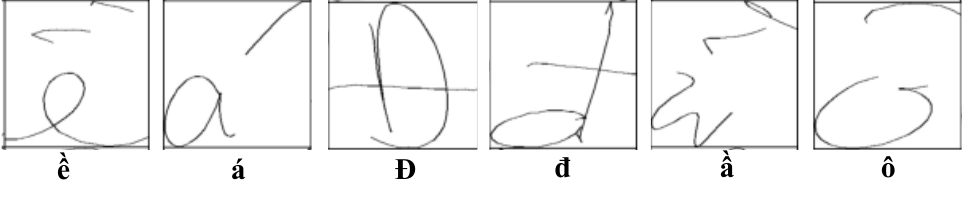}
    \vspace{-0.8cm}
    \caption{Some examples from VNOn-DB dataset.}
    \label{fig:vnondb_examples}
\end{figure}

\vspace{-0.5cm}

The ISI-AIR dataset\cite{rahman2021air} is a publicly available corpus designed for research in mid-air handwriting recognition using motion capture. Collected at the Indian Statistical Institute (ISI), the dataset comprises 3D hand trajectory recordings of English  digits captured using a webcam. Unlike traditional handwriting, ISI-AIR features freehand air gestures without physical contact, introducing challenges such as higher spatial variance, motion blur, and absence of surface constraints. 

    \begin{figure}[h]
    \centering
    \includegraphics[width=1\linewidth]{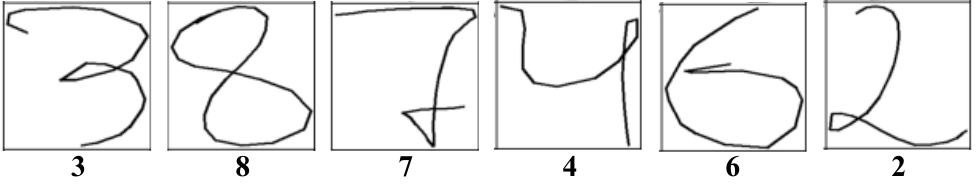}
    \vspace{-0.8cm}
    \caption{Some examples from ISI-Air dataset.}
    \vspace{-0.5cm}
    \label{fig:isi_examples}
\end{figure}

\subsection{Implementation Details}
\vspace{-0.1cm}
All experiments are conducted on a single NVIDIA RTX A5000 (24 GB) GPU using the PyTorch \cite{ansel2024pytorch2} framework. Training proceeds with the AdamW \cite{loshchilov2018decoupled} optimizer (initial learning rate $1\times10^{-4}$, $\beta_1{=}0.9$, $\beta_2{=}0.999$, weight-decay $0.01$). The learning rate follows a cosine-annealing schedule that decays to zero, and gradients are clipped to an $\ell_{2}$-norm of~1.0. Classification uses cross-entropy with label-smoothing $0.1$, while dropout ($p{=}0.1$) and batch-normalization regularize the stroke pathway. 

We evaluate with overall accuracy \(\textit{Acc}=\tfrac{\sum_{c}TP_c}{N}\) and the class-balanced macro variants of precision, recall and F\(_1\): for each class \(c\) we compute \(P_c=\tfrac{TP_c}{TP_c+FP_c}\), \(R_c=\tfrac{TP_c}{TP_c+FN_c}\) and \(F_{1,c}=\tfrac{2\,TP_c}{2\,TP_c+FP_c+FN_c}\), then average them to obtain \(\overline{P}=\tfrac1C\sum_{c}P_c\), \(\overline{R}=\tfrac1C\sum_{c}R_c\) and \(\overline{F_1}=\tfrac1C\sum_{c}F_{1,c}\). All scores are reported in percentage.

\section{Results}
\definecolor{headergray}{gray}{0.90}

\begin{table}[h]
  \caption{Comparison across datasets. Highest values per metric are highlighted. I: Image, S: Stroke, D: Dual}
  \label{tab:all-dataset-results}
  \centering
  \renewcommand\arraystretch{1.0}
  \setlength\tabcolsep{4pt}
  \begin{tabular}{
        p{3.0cm}           
        S[table-format=1.5]  
        S[table-format=2.1]  
        S[table-format=2.1]  
        S[table-format=2.1]  
        S[table-format=2.1]  
      }
    \toprule
    \textbf{Architecture} &
    {\textbf{Mode}} &
    {\textbf{Acc. (\%)}} &
    {\textbf{Precision (\%)}} &
    {\textbf{Recall (\%)}} &
    {\textbf{F1 (\%)}} \\
    \midrule
    \rowcolor{headergray}
    \multicolumn{6}{l}{\textbf{IAM Dataset}~\cite{liwicki2005iam}} \\
    \midrule
    HTR-VT~\cite{li2025htr} & I & {\textbf{95.3}} & {\textbf{94.9}} & {\textbf{94.7}} & {\textbf{94.8}} \\
    \textbf{HAT} & I &  {91.5} & {90.8} & {88.0} & {89.4} \\
    \midrule
    LSTM~\cite{greff2016lstm} & S & {\textbf{90.7}} & 89.3 & {\textbf{90.0}} & {\textbf{88.6}} \\
    \textbf{HAT} & S & {89.5} & {\textbf{90.2}} & {85.0} & {87.5} \\
    \midrule
    OLHTR~\cite{xu2024multi} & {D} & {95.3} & {\textbf{95.1}} & {\textbf{94.6}} & {\textbf{93.8}} \\
    \textbf{HAT} & {D} & {\textbf{96.4}} & {94.0} & {92.5} & {93.7} \\
    \midrule
    \rowcolor{headergray}
    \multicolumn{6}{l}{\textbf{VNOn-DB Dataset}~\cite{nguyen2018icfhr}} \\
    \midrule
    CNN-LSTM~\cite{le2020end} & I & {\textbf{95.3}} & {\textbf{95.0}} & {\textbf{94.2}} & {\textbf{94.6}} \\
    \textbf{HAT} & I & {92.6} & {91.5} & {90.7} & {91.1} \\
    \midrule
    \textbf{HAT} & S & {72.1} & {71.0} & {68.5} & {69.7} \\
    \midrule
    \textbf{HAT} & {D} & {\textbf{95.8}} & {\textbf{95.5}} & {\textbf{95.0}} & {\textbf{95.2}} \\
    \midrule
    \rowcolor{headergray}
    \multicolumn{6}{l}{\textbf{ISI-Air Dataset}~\cite{rahman2021air}} \\
    \midrule
    \textbf{HAT} & I & {99.5} & {98.7} & {99.1} & {98.4} \\
    \midrule
    \textbf{HAT} & S & {98.1} & {97.3} & {97.4} & {98.0} \\
    \midrule
    RNN-LSTM~\cite{rahman2021air} & {D} & {98.7} & {98.6} & {98.5} & {98.6} \\
    \textbf{HAT} & {D} & {\textbf{99.8}} & {\textbf{99.5}} & {\textbf{98.2}} & {\textbf{98.7}} \\
    \bottomrule
  \end{tabular}

  \vspace{4pt}%
\end{table}
\begin{figure}[H]
    \centering
    \includegraphics[width=1\linewidth]{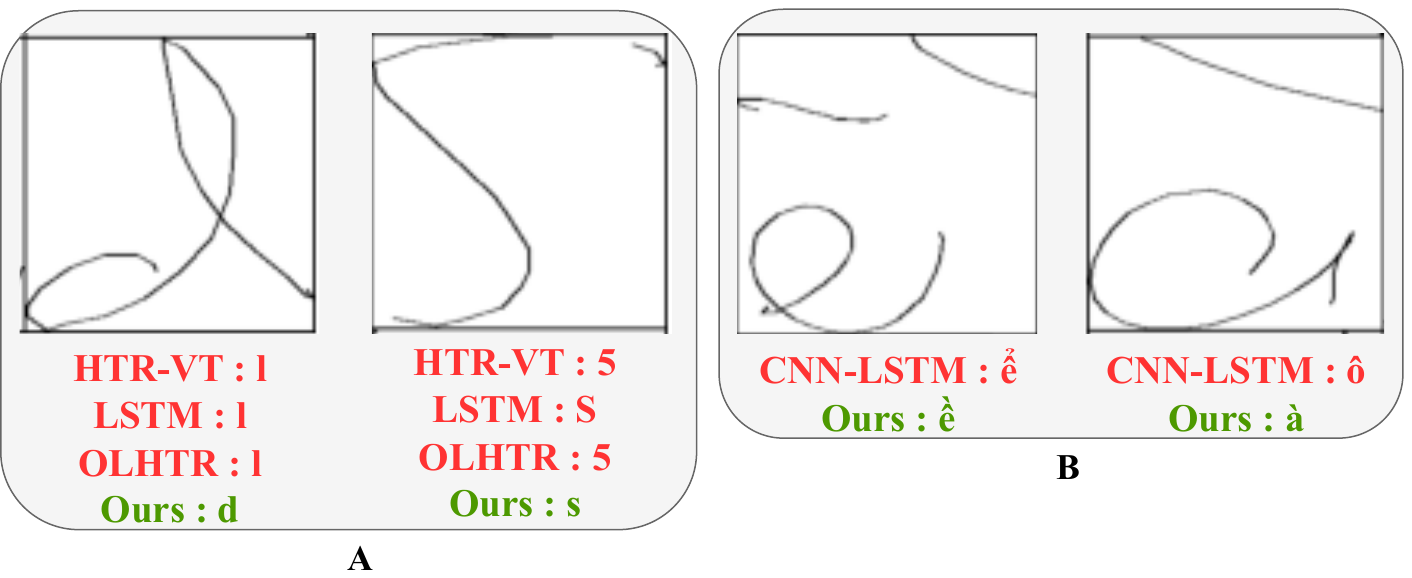}
    \vspace{-4.8cm}
    \caption{Qualitative results (A): IAM-OnDB, (B): Vn-OnDB. Red denotes incorrect recognition, our methods show correctly recognized class in green for all cases shown here.}
    \label{fig:failure_cases}
\end{figure}

Table~\ref{tab:all-dataset-results} displays the performance of comparable models in literature against our proposed recognizer. For the \textbf{IAMOn-DB} dataset, we note a mean 1.5\% improvement in acurracy across all modes (image-only, stroke-only and dual input). OLHTR \cite{xu2024multi} exceeds for dual input marginally (\(<1\%)\) in precision, recall and F-1 score. We observe an interesting 3.8\% gain in accuracy for image-only mode, highlighting the robustness and utility of our framework even when both inputs are not available. 

For \textbf{VNOn-DB} and  we provide their first dual-input benchmarking results. For VNOn-DB, we note an accuracy of 95.8\% for dual input, as opposed to 72.1\% on stroke-only mode. This further reinforces our claim, because although we assume the per-character stroke richness and vocabulary size (145) of the dataset would be much higher than IAMOn-DB, simply relying on stroke information leads to confusion in predictions. Combining image inputs and correlating offline characteristics with the strokes leads to robust identification. 

We notice a similar trend in accuracy for the \textbf{ISI-Air} dataset, with our dual-input model achieving a 1.7\% increase from stroke-only input mode, and an overall 1.1\% increase compared to the traditional RNN-LSTM architecture previously reported in literature \cite{rahman2021air}.

We display some qualitative recognition results in Figure \ref{fig:failure_cases}. It is interesting to note that even though character isolation removes the context of the word it is taken from, our model provides correct results for slightly varying characters even when other paradigms fail. 
\subsection{Ablations}

We provide insights into various settings of feature extraction by swapping out the patch encoder backbone shown in Table \ref{tab:feature_extractor_study}. Swin-B 224~\cite{liu2021swin} displays the best accuracy when used in training; by using frozen weights, the accuracy drops notably by \(10\%\). At its lightest setting (ResNet-18)~\cite{he2016deep} we get an accuracy of 79.13\% highlighting the impact of our chosen Swin-B~\cite{liu2021swin} backbone. Training cost is considerably alleviated since our model converges at the 4th epoch itself for the chosen setting.

\newcommand{\frozenIcon}{\textcolor{cyan}{\faSnowflake[regular]}}
\newcommand{\trainIcon}{\textcolor{orange}{\faFire}}
\begin{table}[h]
  \caption{Ablation study with different image feature extractors.}
  \label{tab:feature_extractor_study}
  \centering
  \renewcommand{\arraystretch}{1.0}
  \setlength\tabcolsep{4pt}
  \begin{tabular}{%
      p{3.0cm}      
      c             
      r             
      r             
      r}            
    \toprule
    \multicolumn{1}{l}{\textbf{Feature Extractor}} &
    \multicolumn{1}{c}{\textbf{Status}} &
    \multicolumn{1}{c}{\textbf{Acc.\ (\%)}} &
    \multicolumn{1}{c}{\textbf{\#Params (M)}} &
    \multicolumn{1}{c}{\textbf{Conv.\ Epochs}}\\
    \midrule
    \shortstack[l]{ResNet-18\cite{he2016deep}} & \frozenIcon  & 79.13 & {--} & 15th \\
    \shortstack[l]{ResNet-18\cite{he2016deep}} & \trainIcon   & 82.95 & 25.04 & 8th \\
    \shortstack[l]{ResNet-34\cite{he2016deep}} & \frozenIcon  & 76.31 & {--} & 16th \\
    \shortstack[l]{ResNet-34\cite{he2016deep}} & \trainIcon   & 80.23 & 35.14 & 9th \\
    \shortstack[l]{ViT (Base)\cite{dosovitskiy2020image}} & \frozenIcon & 82.42 & {--} & 9th \\
    \shortstack[l]{ViT (Base)\cite{dosovitskiy2020image}} & \trainIcon  & 86.14 & 93.47 & 4th \\
    \shortstack[l]{Swin-B 224\cite{liu2021swin}} & \frozenIcon & 82.33 & {--} & 22nd \\
    \shortstack[l]{\textbf{Swin-B 224}\cite{liu2021swin}} & \textbf{\trainIcon} & \textbf{92.42} & 94.48 & 4th \\
    \bottomrule
  \end{tabular}

  \vspace{3pt}
  \footnotesize{All results are measured on the IAMOn-DB for dual inputs. \#Params refers to the parameters of the complete model, including both the stroke and the image branch. \frozenIcon \ denotes frozen and \trainIcon \ denotes trainable parameters}
\end{table}



\begin{figure*}[t]
  \centering
  \begin{minipage}[t]{0.60\textwidth}
    \captionsetup{type=table}
    \captionof{table}{Comparison of convergence with other models on IAMOn-DB~\cite{liwicki2005iam}}
    \label{tab:convergence_study_iam}
    \small
    \setlength\tabcolsep{4pt}
    \renewcommand{\arraystretch}{1.05}

    \begin{adjustbox}{max width=\textwidth}
      \begin{tabular}{l c c c}
        \toprule
        \textbf{Models} &
        \textbf{Acc.\,(\%)} &
        \textbf{\#Params (M)} &
        \textbf{Conv.\ Epochs} \\
        \midrule
        LSTM~\cite{greff2016lstm}          & 90.07 & \textbf{--}   & 150  \\
        \textbf{HAT (Strokes)}             & 89.50 &  4.3 &  46  \\
        \midrule
        HTR-VT~\cite{li2025htr}            & 95.30 & 53.5 & 1\,165 \\
        \textbf{HAT (Images)}              & 91.50 & 88.6 &   4  \\
        \midrule
        OLHTR~\cite{xu2024multi}           & 95.30 & \textbf{--}   & \textbf{--}   \\
        \textbf{HAT (Fusion)}          & 96.42 & 94.4 &  3  \\
        \bottomrule
      \end{tabular}
    \end{adjustbox}

    \vspace{2pt}
    \footnotesize “\#Params” denotes total model parameters; “--”\,= not
    reported in the paper.
  \end{minipage}%
  \hfill
  \begin{minipage}[t]{0.36\textwidth}
    \vspace{0pt} 
    We benchmark the proposed HAT variants against representative
    on-line/off-line recognisers.  
    We observe LSTM and recent transformer-based systems require hundreds to thousands of epochs.  
    In contrast, the efficient-fusion HAT converges in just
    \textbf{three} epochs while retaining competitive accuracy
    (see Table~\ref{tab:convergence_study_iam}).
  \end{minipage}
\end{figure*}

\begin{figure*}[h]
  \centering
  \begin{minipage}[t]{0.47\textwidth}   
    \vspace{0pt}                        
    We benchmark two fusion strategies for our HAT architecture on
    IAMOn-DB. Early and mid-level fusions are contrasted with the
    transformer-based OLHTR baseline.  
    Numerical results are summarised in
    Table~\ref{tab:fusion_study}.
  \end{minipage}%
  \hfill
  \begin{minipage}[t]{0.50\textwidth}   
    \captionsetup{type=table}
    \captionof{table}{Fusion-level comparison of our models on
      IAMOn-DB~\cite{liwicki2005iam}.}
    \label{tab:fusion_study}
    \small
    \setlength\tabcolsep{4pt}
    \renewcommand{\arraystretch}{1.05}

    \begin{adjustbox}{max width=\textwidth}
      \begin{tabular}{l c c}
        \toprule
        \textbf{Model Variant} & \textbf{Fusion} & \textbf{Acc.\,(\%)} \\
        \midrule
        HAT & Early & \textbf{96.40} \\
        HAT & Middle & 92.10 \\
        OLHTR~\cite{xu2024multi} & Late & 95.30 \\
        \bottomrule
      \end{tabular}
    \end{adjustbox}

    \vspace{2pt}
    \footnotesize
    “Acc.” = accuracy.
  \end{minipage}
\end{figure*}

\begin{figure*}[t]
  \centering
  \begin{minipage}[t]{0.50\textwidth}         
    \captionsetup{type=table}
    \captionof{table}{Robustness of the dual-trained model to modality
      dropout (IAMOn-DB).  $\Delta$ is the absolute accuracy drop
      relative to the full-input baseline.}
    \label{tab:modality_dropout}
    \small
    \setlength\tabcolsep{6pt}
    \renewcommand{\arraystretch}{1.08}

    \begin{adjustbox}{max width=\textwidth}
      \begin{tabular}{
          l                    
          S[table-format=2.1]  
          S[table-format=+2.1] 
        }
        \toprule
        \textbf{Train $\rightarrow$ Test} &
        {\textbf{Acc.\,(\%)}} &
        {\textbf{$\Delta$\,(\%)}} \\
        \midrule
        Dual $\rightarrow$ Dual        & 92.4 &  0.0 \\
        Dual $\rightarrow$ Image-only  & 88.1 & -4.3 \\
        Dual $\rightarrow$ Stroke-only & 85.7 & -6.7 \\
        \bottomrule
      \end{tabular}
    \end{adjustbox}
  \end{minipage}%
  \hfill
  \begin{minipage}[t]{0.45\textwidth}        
    \vspace{0pt}  
    We train a single model with dual modalities and then evaluate it under three conditions:
    (i) the full input, (ii) image stream only and
    (iii) stroke stream only .  
    The network gracefully degrades
    when a modality is absent at test time.  A modest drop of 4–7\%
    shows the model remains fairly robust to real-world sensor failure.
  \end{minipage}
\end{figure*}

\section{Discussion}
\label{sec:discussion}

\begin{figure}[H]
    \centering
    \includegraphics[width=6cm, height=4cm]{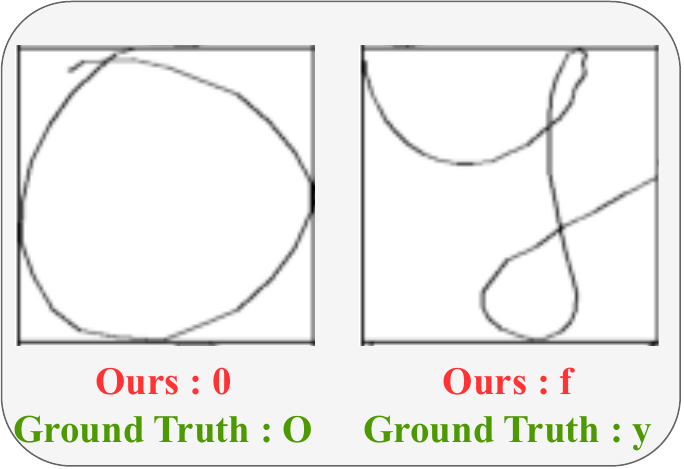}
    \caption{Failure cases for our model.}
    \label{fig:wrong predictions}
\end{figure}

\textbf{Limitations.}  
The current system recognises isolated glyphs, assumes pre-segmented
inputs, and depends on a training visual backbone of 90 M parameters. 
Performance has not been audited across writer demographics or fine‐grained
stroke disorders, and segmentation errors in IAM0n-DB and VNOn-DB introduce small but
uncorrected label noise. We also notice some failure cases for our model in Figure \ref{fig:wrong predictions}.

\textbf{Future Work.}  
Future work lies in curation of a large level multilingual word level and line level dataset to extend the usage of this framework to real applications. There is also potential in providing for manually segmented character information that might help in better benchmarking results.

\textbf{Ethical Considerations.}  
Pen trajectories may act as a biometric, enabling unintended
writer-identification.  Some security concerns might stem from learned correlations between motor patterns through stroke and subsequent formed character images, which warrants careful auditing of software that uses this framework.

\section{Conclusion}

We propose HATCharClassifier, a novel framework that utilises early fusion utilising online and offline input modalities for handwritten text recognition. Our method displays the utility of capturing the correlation between the two modalities using cross-modal querying, leading to more robust recognition across multiple datasets as discussed above. Our approach opens a new direction in this field, motivating future work for word-level and line-level air-writing dataset collection and benchmarking. We notice some limitations such as minor sparse inconsistencies in character stroke segmentation used in preprocessing for the IAMOn-DB and VNOn-DB datasets from word-level, and heavier parameter and floating-point operations (FLOPs) than other existent methods using dual input. We believe this leaves potential for further improvement in future work.

\bibliographystyle{splncs04}
\bibliography{main}

\end{document}